\documentclass[10pt,twocolumn,letterpaper]{article}

\usepackage[pagenumbers]{iccv} 

\definecolor{iccvblue}{rgb}{0.21,0.49,0.74}
\usepackage[pagebackref,breaklinks,colorlinks,allcolors=iccvblue]{hyperref}
\usepackage{graphicx}
\usepackage{amsmath}
\usepackage{amssymb}
\usepackage{booktabs}
\usepackage{float}

\title{DreamActor-M1: Holistic, Expressive and Robust Human Image Animation\\ with Hybrid Guidance}

\author{
Yuxuan Luo\thanks{Equal Contribution.} \quad Zhengkun Rong\footnotemark[1] \quad Lizhen Wang\footnotemark[1] \quad Longhao Zhang\footnotemark[1] \quad Tianshu Hu\footnotemark[1] \thanks{Corresponding author.} \quad  Yongming Zhu \\
Bytedance Intelligent Creation\\
{\tt\small \{luoyuxuan, rongzhengkun, wanglizhen.2024, zhanglonghao.zlh, tianshu.hu, } \\
{\tt\small zhuyongming\}@bytedance.com}
}

\begin{document}

\twocolumn[{
\maketitle
\begin{figure}[H]
\hsize=\textwidth
\centering
\vspace{-1.0cm}
\includegraphics[width=2.1\linewidth]{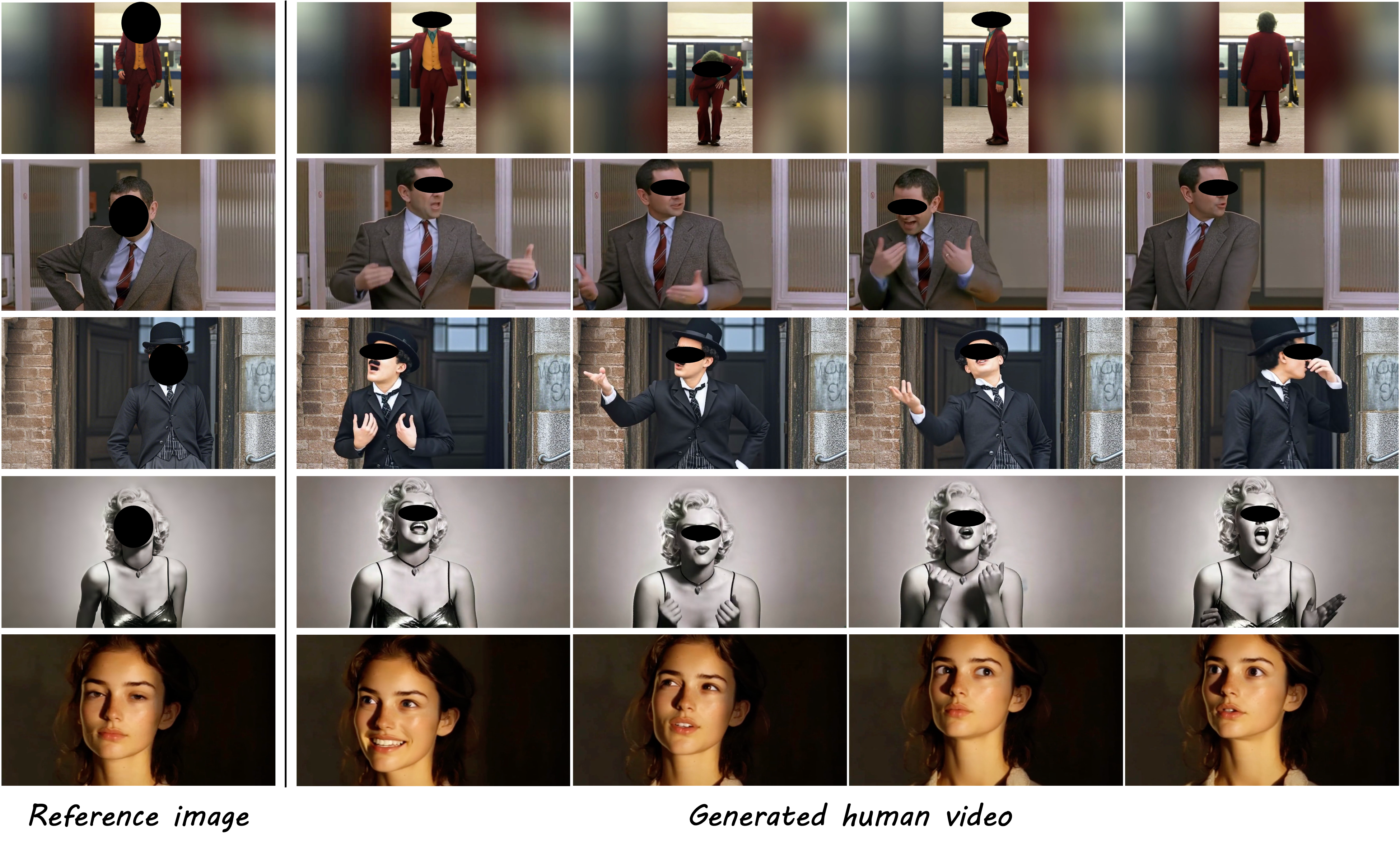}
\caption{We introduce DreamActor-M1, a DiT-based human animation framework, with hybrid guidance to achieve fine-grained holistic controllability, multi-scale adaptability, and long-term temporal coherence.}
\label{fig:teaser}
\end{figure}
}]
{
  \renewcommand{\thefootnote}%
    {\fnsymbol{footnote}}
  \footnotetext[1]{Equal contribution.}
  \footnotetext[2]{Corresponding author.}
}

\begin{abstract}
While recent image-based human animation methods achieve realistic body and facial motion synthesis, critical gaps remain in fine-grained holistic controllability, multi-scale adaptability, and long-term temporal coherence, which leads to their lower expressiveness and robustness. We propose a diffusion transformer (DiT) based framework, DreamActor-M1, with hybrid guidance to overcome these limitations. 
For motion guidance, our hybrid control signals that integrate implicit facial representations, 3D head spheres, and 3D body skeletons achieve robust control of facial expressions and body movements, while producing expressive and identity-preserving animations.
For scale adaptation, to handle various body poses and image scales ranging from portraits to full-body views, we employ a progressive training strategy using data with varying resolutions and scales.
For appearance guidance, we integrate motion patterns from sequential frames with complementary visual references, ensuring long-term temporal coherence for unseen regions during complex movements.
Experiments demonstrate that our method outperforms the state-of-the-art works, delivering expressive results for portraits, upper-body, and full-body generation with robust long-term consistency. Project Page: \href{https://grisoon.github.io/DreamActor-M1/}{\emph{https://grisoon.github.io/DreamActor-M1/}}.
\end{abstract}    
\section{Introduction}
\label{sec:intro}


Human image animation has become a hot research direction in video generation and provides potential applications for film production, the advertising industry, and video games. While recent advances in image and video diffusion have enabled basic human motion synthesis from a single image, previous works like~\cite{hu2024animate, AnimateX2025, zhu2024champ, li2024dispose, zhang2024mimicmotion, xu2024magicanimate, peng2024controlnext} have made great progress in this area. However, existing methods remain constrained to coarse-grained animation. There are still some critical challenges in achieving fine-grained holistic control (e.g., subtle eye blinks and lip tremors), generalization ability to multi-scale inputs (portrait/upper-body/full-body), and long-term temporal coherence (e.g. long-term consistency for unseen garment areas). In order to handle these complex scenarios, we propose a DiT-based framework, DreamActor-M1, to achieve holistic, expressive and robust human image animation with hybrid guidance.
Recent developments in image-based animation have explored various directions, yet critical shortcomings persist in attaining photorealistic, expressive, and adaptable generation for practical applications. While single-image facial animation approaches employ landmark-driven or 3DMM-driven methodologies through GAN~\cite{siarohin2019first, zhao2022thin, wang2021one, sun2023next3d, deng2024portrait4d, wang2023rodin} or NeRF~\cite{yu2023nofa} for expression manipulation, they frequently face limitations in image resolution and expression accuracy.
Although certain studies attain robust expression precision via latent face representations~\cite{drobyshev2022megaportraits, drobyshev2024emoportraits, xu2025vasa, wang2022pdfgc, zhu2024infp} or achieve enhanced visual quality with diffusion-based architectures~\cite{tian2024emo,X-NeMo2025}, their applicability is typically restricted to portrait regions, failing to meet the broader demands of real-world scenarios.
Diffusion-based human image animation~\cite{hu2024animate,xu2024magicanimate,chang2023magicpose,zhang2024mimicmotion,tu2024stableanimator} are able to generate basic limb articulation, plausible garment deformations and hair dynamics, but neglect the fine-grained facial expressions reenactment.
Current solutions remain under-explored in addressing holistic control of facial expressions and body movements while failing to accommodate real-world multi-scale deployment requirements. Furthermore, existing methods fail to maintain temporal coherence in long-form video synthesis, especially for unseen areas in the reference image.

In this work, we focus on addressing multi-scale driven synthesis, fine-grained face and body control, and long-term temporal consistency for unseen areas. Solving these challenges is non-trivial. 
First, it is very challenging to accurately manage both detailed facial expressions and body movements using a single control signal, especially for the precise control over subtle facial expressions.
Second, due to incomplete input data and the inability to generate extended video sequences in a single pass, the model inevitably loses information about unseen regions (such as clothing textures on the back) during the continuation process that relies solely on the reference image and the final frame of prior clips. This gradual information decay leads to inconsistencies in unseen areas across sequentially generated segments.
Third, under multi-scale inputs, the varying information density and focus priorities make it hard to achieve holistic and expressive animation within a single framework.
To tackle these issues, we introduce stronger hybrid control signals, and complementary appearance guidance that can fill missing information gaps, and a progressive training strategy with a training dataset including diverse samples at different scales (e.g. portrait talking and full-body dancing).


Specifically, for motion guidance, we design a hybrid control signals including implicit face latent for fine-grained control of facial expressions, explicit head spheres for head scale and rotation and 3D body skeletons for torso movements and bone length adjustment which achieves robust adaptation across substantial shape variations.
For scenarios with limited information (e.g., multi-turn rotations or partial-body references), we introduce complementary appearance guidance by first sampling distinct poses from target movements, then generating multi-frame references to provide unseen area textures, and finally propagating these references across video segments to maintain consistent details throughout long-term synthesis.
To enable multi-scale adaptation, we train the model with a progressive strategy on a diverse dataset that includes different types of scenes like portrait acting, upper-body talking, and full-body dancing.
In summary, our key contributions are as follows.

\begin{itemize}
\item We propose a holistic DiT-based framework and a progressive training strategy for human image animation that supports flexible multi-scale synthesis.

\item We design hybrid control signals combining implicit facial representations, explicit 3D head  spheres, and body skeletons to enable the expressive body and facial motion synthesis while supporting diverse character styles.

\item We develop complementary appearance guidance to mitigate information gaps of unseen areas between video segments, enabling consistent video generation over long durations.
\end{itemize}

\section{Related Works}
\label{sec:formatting}

\begin{figure*}
   \centering
   \includegraphics[width=0.99\linewidth]{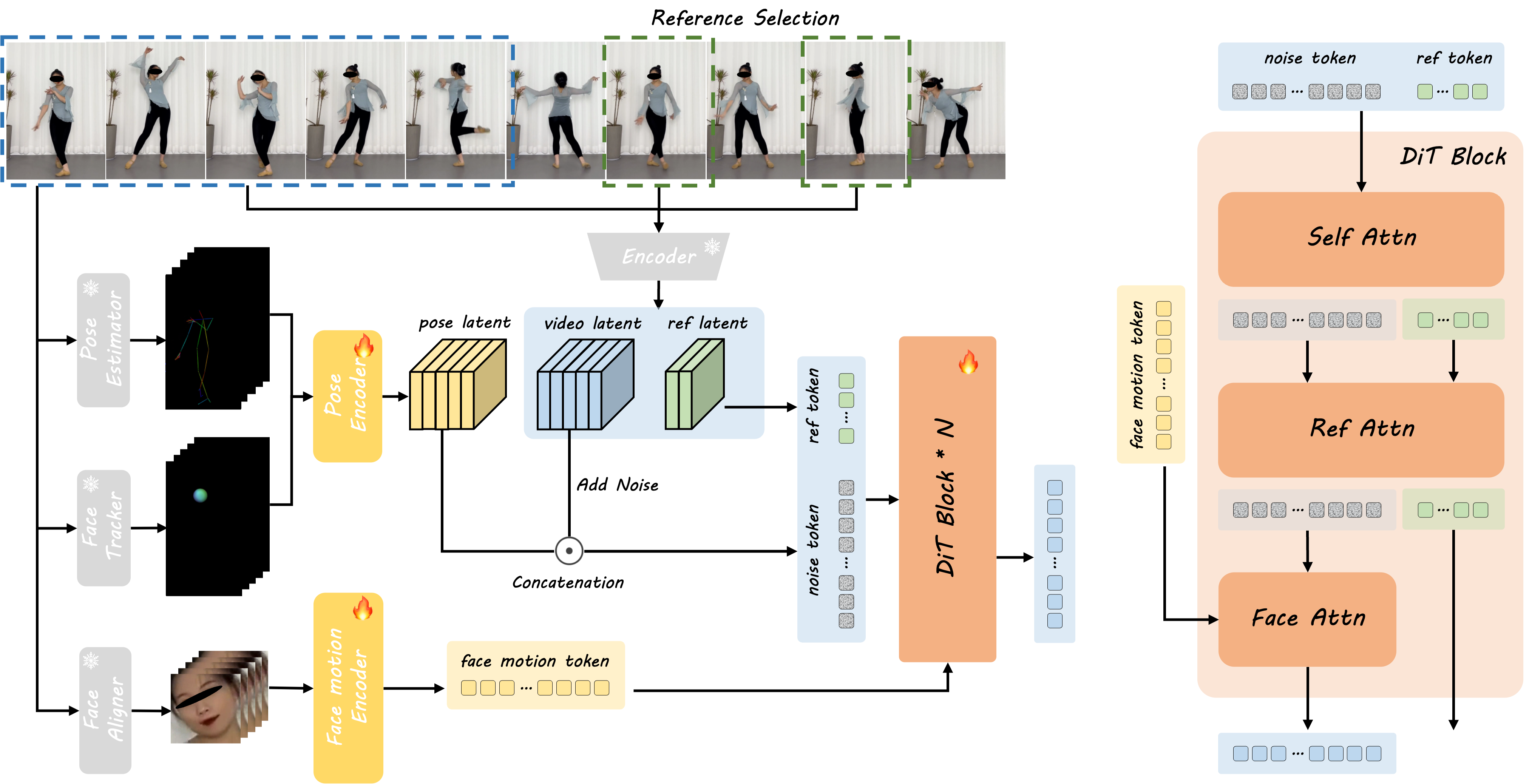}
      \caption{Overview of DreamActor-M1. During the training stage, we first extract body skeletons and head spheres from driving frames and then encode them to the pose latent using the pose encoder. The resultant pose latent is combined with the noised video latent along the channel dimension. The video latent is obtained by encoding a clip from the input full video using 3D VAE. Facial expression is additionally encoded by the face motion encoder, to generate implicit facial representations. Note that the reference image can be one or multiple frames sampled from the input video to provide additional appearance details during training and the reference token branch shares weights of our DiT model with the noise token branch. Finally, the denoised video latent is supervised by the encoded video latent. Within each DiT block, the face motion token is integrated into the noise token branch via cross-attention (Face Attn), while appearance information of ref token is injected to noise token through concatenated self-attention (Self Attn) and subsequent cross-attention (Ref Attn).}
\vspace{-0.3cm}
   \label{fig:overview}
\end{figure*}

Recent advancements in human image animation can be broadly categorized into single-image facial animation and body animation, each addressing distinct technical challenges in photorealistic and expressive human motion synthesis.

\subsection{Single-Image Facial Animation}


Early work primarily relied on GAN to create portrait animation through warping and rendering. These studies typically focused on improving driving expressiveness by exploring various motion representations including neural keypoints~\cite{siarohin2019first,zhao2022thin,guo2024liveportrait,wang2021one} or 3D face model parameters~\cite{ren2021pirenderer,Doukas_2021_ICCV}. These methods can effectively decouple identity and expression features, but the reproduction of expressions, especially subtle and exaggerated ones, is quite limited. Another class of methods learned latent representation~\cite{wang2022pdfgc,drobyshev2022megaportraits,drobyshev2024emoportraits} directly from the driving face, enabling higher-quality expression reproduction. However, limited by GAN, these methods faced challenges in generating high-quality results and adapting to different portrait styles.
In recent years, diffusion-based methods have demonstrated strong generative capabilities, leading to significant advancements in subsequent research.
EMO~\cite{tian2024emo} first introduced the ReferenceNet into the diffusion-based portrait video generation task.
Follow-your-emoji \cite{ma2024followemoji} utilized expression-aware landmarks with facial fine-grained loss for precise motion alignment and micro-expression details.
Megactor-${\Sigma}$~\cite{yang2024megactor} designed a diffusion transformer integrating audio-visual signals for multi-modal control.
X-Portrait~\cite{xie2024x} combined ControlNet for head pose and expression with patch-based local control and cross-identity training.
X-Nemo~\cite{X-NeMo2025} developed 1-D latent descriptors to disentangle identity-motion entanglement.
SkyReels-A1~\cite{qiu2025skyreels} proposed a portrait animation framework leveraging the diffusion transformer~\cite{peebles2023scalable}.

\subsection{Single-Image Body Animation}

As a pioneering work in body animation, MRAA~\cite{siarohin2021motion} proposed distinct motion representations for animating articulated objects in an unsupervised manner.
Recent advancements in latent diffusion models~\cite{rombach2022high} have significantly boosted the development of body animation~\cite{ma2024followemoji, chang2023magicpose, musepose, wang2024unianimate, karras2023dreampose, wang2024disco}.
Animate Anyone~\cite{hu2024animate} introduced an additional ReferenceNet to extract appearance features from reference images.
MimicMotion~\cite{zhang2024mimicmotion} designed a confidence-aware pose guidance mechanism and used facial landmarks to reenact facial expressions.
Animate-X~\cite{AnimateX2025} employed implicit and explicit pose indicators for motion and pose features, respectively, achieving better generalization across anthropomorphic characters.
Instead of using skeleton maps, Champ~\cite{zhu2024champ} and Make-Your-Anchor~\cite{huang2024make} leveraged a 3D human parametric model, while MagicAnimate~\cite{xu2024magicanimate} utilized DensePose~\cite{guler2018densepose} to establish dense correspondences.
TALK-Act~\cite{guan2024talk} enhanced the textural awareness with explicit motion guidance to improve the generation quality.
StableAnimator~\cite{tu2024stableanimator} and HIA~\cite{xu2024high} addressed the identity preservation and motion blur modeling, respectively.
Additionally, some previous studies~\cite{yoon2024tpc, li2024dispose, peng2024controlnext} followed a plug-in paradigm, enhancing effectiveness without requiring additional training of existing model parameters.
To address missing appearances under significant viewpoint changes, MSTed~\cite{hong2024free} introduced multiple reference images as input.
MIMO~\cite{men2024mimo} and Animate Anyone 2~\cite{hu2025animate} separately modeled humans, objects, and backgrounds, aiming to animate characters with environmental affordance.
Apart from UNet-based diffusion, recent works~\cite{gan2025humandit, shao2024human4dit} have also begun adapting diffusion transformers~\cite{peebles2023scalable} for human animation.

\section{Method}

\begin{figure*}
   \centering
   \includegraphics[width=0.99\linewidth]{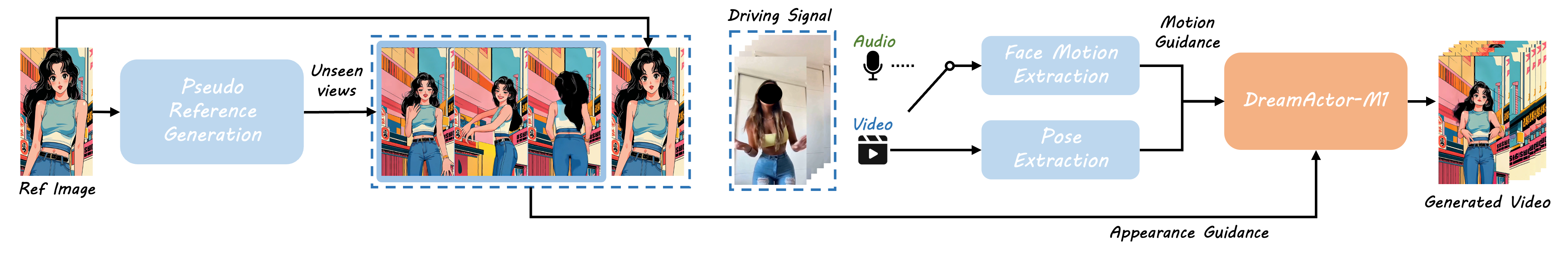}
   \caption{Overview of our inference pipeline. First, we (optionally) generate multiple pseudo-references to provide complementary appearance information. Next, we extract hybrid control signals comprising implicit facial motion and explicit poses (head sphere and body skeleton) from the driving video. Finally, these signals are injected into a DiT model to synthesize animated human videos. Our framework decouples facial motion from body poses, with facial motion signals being alternatively derivable from speech inputs.}
   \label{fig:inf}
\vspace{-0.3cm}
\end{figure*}

Given one or multiple reference images $I_{R}$
along with a driven video $V_{D}$,
our objective is to generate a realistic video that depicts the reference character mimicking the motions present in the driving video. In this section, we begin with a concise introduction to our Diffusion Transformer (DiT) backbone in \cref{sec:3.1}. Following that, we offer an in-depth explanation of our carefully-designed hybrid control signals in \cref{sec:3.2}. Subsequently, we introduce the complementary appearance guidance in \cref{sec:3.3}. Finally, we present the progressive training processes in \cref{sec:3.4}.

\subsection{Preliminaries}\label{sec:3.1}
As shown in \cref{fig:overview}, our overall framework adheres to the Latent Diffusion Model (LDM)~\cite{rombach2022high} for training the model within the latent space of a pre-trained 3D Variational Autoencoder (VAE)~\cite{yu2023language}. We utilize the MMDiT~\cite{esser2024scaling} as the backbone network, which has been pre-trained on text-to-video and image-to-video tasks, Seaweed~\cite{lin2025diffusion}, and follow the pose condition training scheme proposed by OmniHuman-1~\cite{lin2025omnihuman1}. Note that we employ Flow Matching~\cite{lipman2022flow} as the training objective.

Unlike the prevailing ReferenceNet-based human animation approaches~\cite{hu2024animate, hu2025animate}, we refrain from employing a copy of the DiT as the ReferenceNet to inject the reference feature into the DenoisingNet following~\cite{lin2025omnihuman}. 
Instead, we flatten the latent feature $\widetilde{I}_{R}$ and $\widetilde{V}_{D}$ extracted through the VAE, patchify, concatenate them together, and then feed them into the DiT. It facilitates the information interaction between the reference and video frames through 3D self-attention layers and spatial cross-attention layers integrated throughout the entire model. Specifically, in each DiT block, given the concatenated token $T \in \mathbb{R}^{(t \times h \times w) \times c}$, we perform self-attention along the first dimension. Then, we split it into $T_{R} \in \mathbb{R}^{t_{R} \times (h \times w) \times c}$ and $T_{D} \in \mathbb{R}^{t_{D} \times (h \times w) \times c}$ and reshape them to $T_{R} \in \mathbb{R}^{1 \times (h \times w \times t_{R}) \times c}$ and $T_{D} \in \mathbb{R}^{t_{D} \times (h \times w) \times c}$ to perform cross-attention along the second dimension.

\subsection{Hybrid Motion Guidance}\label{sec:3.2}
To achieve expressive and robust human animation, in this paper, we intricately craft the motion guidance and propose hybrid control signals comprising implicit facial representations, 3D head spheres, and 3D body skeletons.

\noindent\textbf{Implicit Facial Representations.}
In contrast to conventional approaches that rely on facial landmarks for expression generation, our method introduces implicit facial representations. This innovative approach not only enhances the preservation of intricate facial expression details but also facilitates the effective decoupling of facial expressions, identity and  head pose, enabling more flexible and realistic animation.
Specifically, our pipeline begins by detecting and cropping the faces in the driving video, which are then resized to a standardized format $F \in \mathbb{R}^{t \times 3 \times 224 \times 224}$. A pre-trained face motion encoder $\textbf{E}_{f}$ and an MLP layer are employed to encode faces to face motion tokens $M \in \mathbb{R}^{t \times c}$. The $M$ is fed into the DiT block through a cross-attention layer. 
The face motion encoder is initialized using an off-the-shelf facial representation learning method~\cite{wang2022pdfgc}, which has been pre-trained on large-scale datasets to extract identity-independent expression features. This initialization not only accelerates convergence but also ensures that the encoded motion tokens are robust to variations in identity, focusing solely on the nuances of facial expressions. By leveraging implicit facial representations, our method achieves superior performance in capturing subtle expression dynamics while maintaining a high degree of flexibility for downstream tasks such as expression transfer and reenactment.
It is worth noting that we additionally train an audio-driven encoder capable of mapping speech signals to face motion token. This encoder enables facial expression editing, particularly lip-syncing, without the need for a driving video. 

\noindent\textbf{3D Head Spheres.}
Since the implicit facial representations are designed to exclusively control facial expressions, we introduce an additional 3D head sphere to independently manage head pose. This dual-control strategy ensures that facial expressions and head movements are decoupled, enabling more precise and flexible animation. Specifically, we utilize an off-the-shelf face tracking method~\cite{wang2022faceverse} to extract 3D facial parameters from the driving video, including camera parameters and rotation angles. These parameters are then used to render the head as a color sphere projected onto the 2D image plane. The sphere’s position is carefully aligned with the position of the driving head in the video frame, ensuring spatial consistency. Additionally, the size of the sphere is scaled to match the reference head’s size, while its color is dynamically determined by the driving head’s orientation, providing a visual cue for head rotation. This 3D sphere representation offers a highly flexible and intuitive way to control head pose, significantly reducing the model’s learning complexity by abstracting complex 3D head movements into a simple yet effective 2D representation. This approach is particularly advantageous for preserving the unique head structures of reference characters, especially those from anime and cartoon domains.

\noindent\textbf{3D Body Skeletons.}
For body control, we introduce 3D body skeletons with bone length adjustment. In particular, we first use 4DHumans~\cite{goel2023humans} and HaMeR~\cite{pavlakos2024reconstructing} to estimate body and hand parameters of the SMPL-X~\cite{SMPL-X:2019} model. Then we select the body joints, project them onto the 2D image plane, and connect them with lines to construct the skeleton maps. 
We opt to use skeletons instead of rendering the full body, as done in Champ~\cite{zhu2024champ}, to avoid providing the model with strong guidance on body shape. By leveraging skeletons, we encourage the model to learn the shape and appearance of the character directly from the reference images. This approach not only reduces bias introduced by predefined body shapes but also enhances the model's ability to generalize across diverse body types and poses, leading to more flexible and realistic results.
The body skeletons and the head spheres are concatenated in the channel dimension, and fed into a pose encoder $\textbf{E}_{p}$ to obtain the pose feature. The pose feature and noised video feature are then concatenated and processed by an MLP layer to obtain the noise token.

During inference, to address variations in skeletal proportions across subjects, we adopt a normalization process to adjust the bone length. First, we use a pre-trained image editing model~\cite{shi2024seededit} to transform reference and driving images into a standardized A-pose configuration. Next, we leverage RTMPose~\cite{jiang2023rtmpose} to calculate the skeletal proportions of both the driving subject and the reference subject. Finally, we perform anatomical alignment by proportionally adjusting the bone lengths of the driving subject to match the skeletal measurements of the reference subject.

\begin{figure*}
   \centering
   \includegraphics[width=0.99\linewidth]{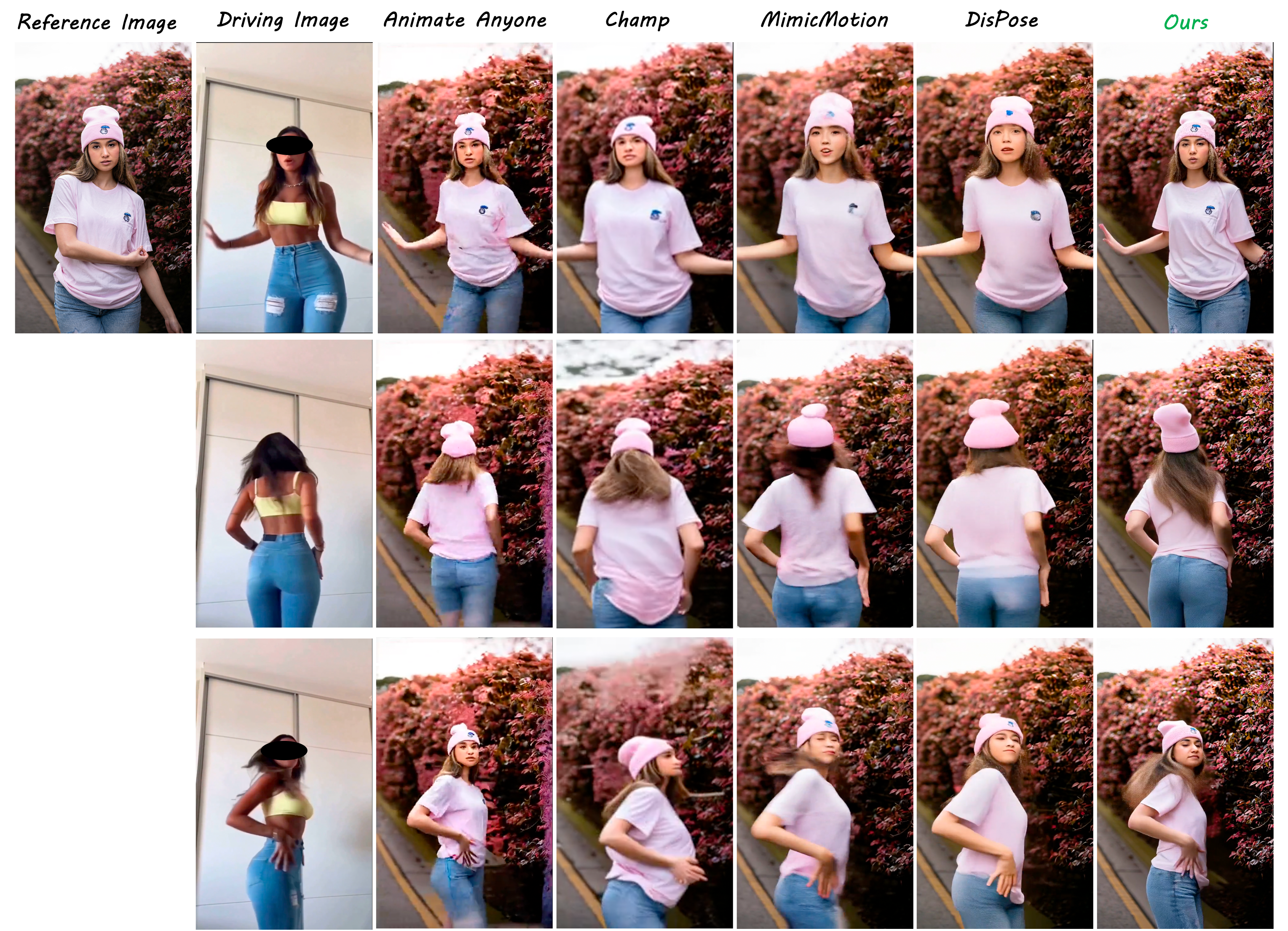}
   \caption{The comparisons with human image animation works, Animate Anyone~\cite{hu2024animate}, Champ~\cite{zhu2024champ}, MimicMotion~\cite{zhang2024mimicmotion} and DisPose~\cite{li2024dispose}. Our method demonstrates results with better fine-grained motions, identity preservation, temporal consistency and high fidelity.}
   \label{fig:4-qualitative-pose-v1}
\vspace{-0.3cm}
\end{figure*}

\subsection{Complementary Appearance Guidance}\label{sec:3.3}

We propose a novel multi-reference injection protocol to enhance the model's capability for robust multi-scale, multi-view, and long-term video generation. This approach addresses the challenges of maintaining temporal consistency and visual fidelity across diverse viewing angles and extended timeframes. During training, we compute the rotation angles for all frames in the input video and sort them based on their z-axis rotation values (yaw). From this sorted set, we strategically select three key frames corresponding to the maximum, minimum, and median z-axis rotation angles. These frames serve as representative viewpoints, ensuring comprehensive coverage of the object's orientation. Furthermore, for videos featuring full-body compositions, we introduce an additional step: a single frame is randomly selected and cropped to a half-body portrait format, which is then included as an auxiliary reference frame. This step enriches the model's understanding of both global and local structural details.

During inference, our protocol offers an optional two-stage generation mode to handle challenging scenarios, such as cases where the reference image is a single frontal half-body portrait while the driving video features full-body frames with complex motions like turning or side views. First, the  model is utilized to synthesize a multi-view video sequence from the single reference image. This initial output captures a range of plausible viewpoints and serves as a foundation for further refinement.We apply the same frame selection strategy used during training to select the most informative frames. These selected frames are then reintegrated into the model as the complementary appearance guidance, enabling the generation of a final output that exhibits enhanced spatial and temporal coherence. This iterative approach not only improves the robustness of the model but also ensures high-quality results even under constrained input conditions.

\subsection{Progressive Training Process}\label{sec:3.4}
Our training process is divided into three distinct stages to ensure a gradual and effective adaptation of the model. In the first stage, we utilize only two control signals: 3D body skeletons and 3D head spheres, deliberately excluding the implicit facial representations. This initial stage is designed to facilitate the transition of the base video generation model to the task of human animation. By avoiding overly complex control signals that may hinder the model's learning process, we allow the model to establish a strong foundational understanding of the task. In the second stage, we introduce the implicit facial representations while keeping all other model parameters frozen. During this stage, only the face motion encoder and face attention layers are trained, enabling the model to focus on learning the intricate details of facial expressions without the interference of other variables. Finally, in the third stage, we unfreeze all model parameters and conduct a comprehensive training session, allowing the model to fine-tune its performance by jointly optimizing all components. This staged approach ensures a robust and stable training process, ultimately leading to a more effective and adaptable model.

\section{Experiments}

\begin{figure*}
   \centering
   \includegraphics[width=0.99\linewidth]{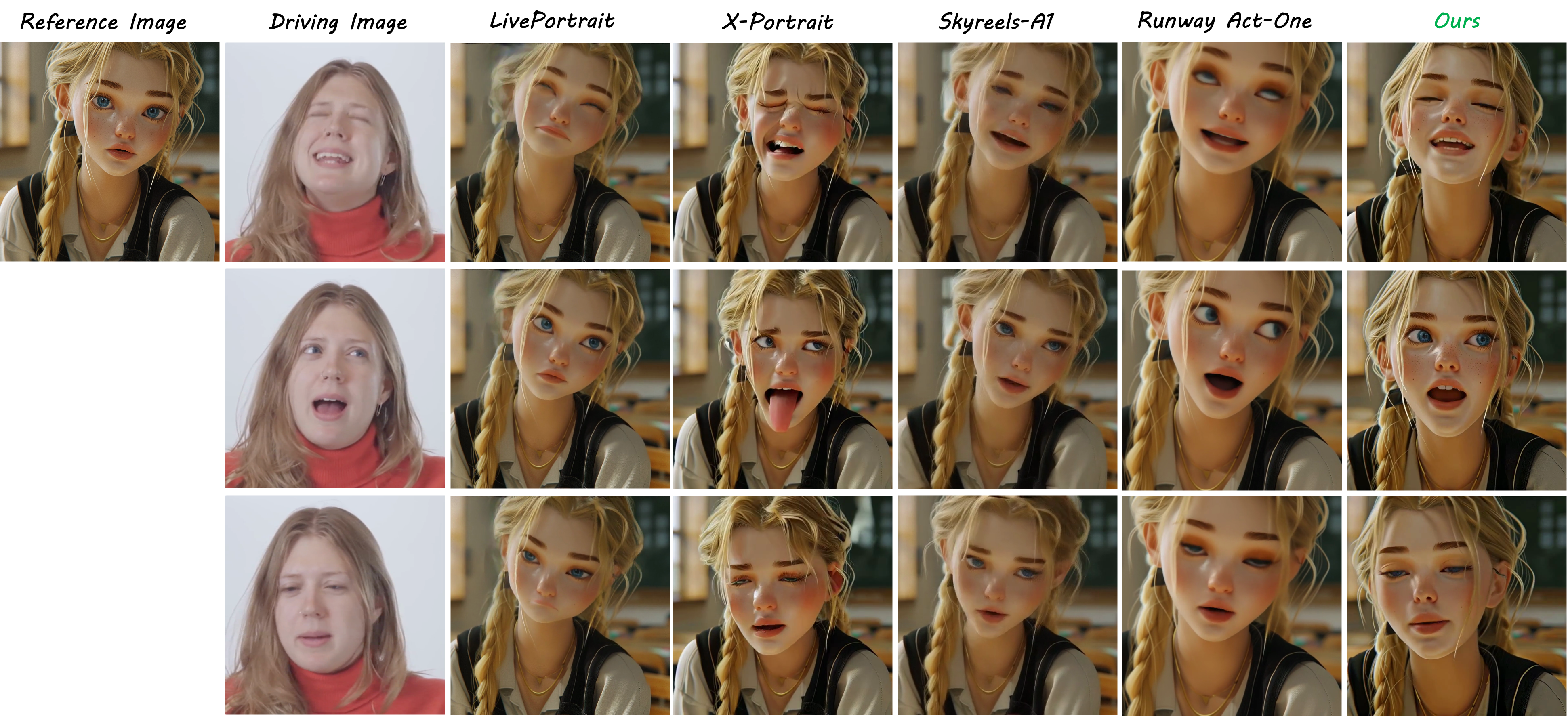}
   \caption{Our comparisons with portrait image animation works, LivePortrait~\cite{guo2024liveportrait}, X-Portrait~\cite{xie2024x}, Skyreels-A1~\cite{qiu2025skyreels} and Runway Act-One~\cite{runwayactone}. Our method demonstrates more accurate and expressive portrait animation capabilities.}
   \label{fig:4-qualitative-portrait-v1}
\vspace{-0.3cm}
\end{figure*}

\subsection{Experimental Setups}\label{sec:4.1}

\noindent\textbf{Implementation Details.}
Our training weights are initialized from a pretrained image-to-video DiT model~\cite{lin2025diffusion} and warm up with a condition training strategy~\cite{lin2025omnihuman1}. Then, we train the first stage with 20,000 steps, the second stage with 20,000 steps, and the third stage with 30,000 steps. To enhance the model's generalization capability for arbitrary durations and resolutions, during training, the length of the sampled video clips is randomly selected from 25 to 121 frames, while the spatial resolution is resized to an area of $960\times640$, maintaining the original aspect ratio. All stages are trained with 8 H20 GPUs using the AdamW optimizer with a learning rate of $5e^{-6}$. During inference, each video segment contains 73 frames. To ensure full-video consistency, we use the last latent from the current segment as the initial latent for the next segment, modeling the next segment generation as an image-to-video generation task. The classifier-free guidance (cfg) parameters for both the references and motion control signals are set to 2.5.

\noindent\textbf{Datasets.}
For training, we construct a comprehensive dataset by collecting video data from various sources, totaling 500 hours of footage. This dataset encompasses a diverse range of scenarios, including dancing, sports, film scenes, and speeches, ensuring broad coverage of human motion and expressions. The dataset is balanced in terms of framing, with full-body shots and half-body shots each accounting for approximately 50\% of the data. In addition, we leverage Nersemble~\cite{kirschstein2023nersemble} to further improve the synthesis quality of faces. 
For evaluation, we utilize our collected dataset, which provides a varied and challenging benchmark, enabling a robust assessment of the model’s generalization capabilities across different scenarios.

\noindent\textbf{Evaluation Metrics.}
We adhere to established evaluation metrics employed in prior research, including FID, SSIM, LPIPS, PSNR, and FVD. The first four are used to evaluate the generation quality of each frame, while the last one is used to assess the video fidelity.

\subsection{Comparisons with Existing Methods}\label{sec:4.2}

To comprehensively demonstrate the effectiveness of our work, we conducted experiments on both body animation and portrait animation tasks. 
Note that our method demonstrates strong performance with just a single reference image in most cases. To ensure fairness in comparison with other methods, we only used multiple reference images in the ablation study, while a single reference image was employed in the comparative analysis.
We strongly recommend readers refer to the supplementary video.

\noindent\textbf{Comparisons with Body Animation Methods.}
We perform the qualitative and quantitative evaluation of DreamActor-M1 with our collected dataset and compare with state-of-the-art body animation methods, including Animate Anyone~\cite{hu2024animate},  Champ~\cite{zhu2024champ}, MimicMotion~\cite{zhang2024mimicmotion}, and DisPose~\cite{li2024dispose}, as shown in \cref{tab:ours_body} and \cref{fig:4-qualitative-pose-v1}. We can see that our proposed DreamActor-M1 outperforms the current state-of-the-art results.

\noindent\textbf{Comparisons with Portrait Animation Methods.}
We also compare the DreamActor-M1 with state-of-the-art portrait animation methods, including LivePortrait~\cite{guo2024liveportrait},  X-Portrait~\cite{xie2024x}, SkyReels-A1~\cite{qiu2025skyreels}, and Act-One~\cite{runwayactone}, as shown in \cref{tab:ours_port} and \cref{fig:4-qualitative-portrait-v1}. As shown in \cref{tab:ours_port}, the video-driven results consistently outperform all competing methods across all metrics on our collected dataset.

While facial expressions and head pose are decoupled in our framework, our method can also be extended to audio-driven facial animation. Specifically, we train a face motion encoder to map speech signals to face motion tokens, leading to realistic and lip-sync animations. As an extended application, we omit quantitative comparisons. Please refer to our supplementary video for more results.

\begin{table}
  \centering
  \footnotesize
  \begin{tabular}{l c c c c c c c c c}
    \toprule
     & FID$\downarrow$ & SSIM$\uparrow$ & PSNR$\uparrow$ & LPIPS$\downarrow$ & FVD$\downarrow$
    \\
        \midrule
        AA~\cite{hu2024animate} & 36.72 & 0.791 & 21.74 & 0.266 & 158.3\\
        Champ~\cite{zhu2024champ} & 40.21 & 0.732 & 20.18 & 0.281 & 171.2\\
        MimicMotion~\cite{zhang2024mimicmotion} & 35.90 & 0.799 & 22.25 & 0.253 & 149.9\\
        DisPose~\cite{li2024dispose} & 33.01 & 0.804 & 21.99 & 0.248 & 144.7\\
        \midrule
        Ours & \textbf{27.27} & \textbf{0.821} & \textbf{23.93} & \textbf{0.206} & \textbf{122.0}\\
    \bottomrule
  \end{tabular}
  \caption{Quantitative comparisons with body animation methods on our collected dataset.}
  \label{tab:ours_body}
\vspace{-0.3cm}
\end{table}

\begin{table}
  \centering
  \footnotesize
  \begin{tabular}{l c c c c c c c c c}
    \toprule
     & FID$\downarrow$ & SSIM$\uparrow$ & PSNR$\uparrow$ & LPIPS$\downarrow$ & FVD$\downarrow$
    \\
        \midrule
        LivePortrait~\cite{guo2024liveportrait} & 31.72 & 0.809 & 24.25 & 0.270 & 147.1\\
        X-Portrait~\cite{xie2024x} & 30.09 & 0.774 & 22.98 & 0.281 & 150.9\\
        SkyReels-A1~\cite{qiu2025skyreels} & 30.66 & 0.811 & 24.11 & 0.262 & 133.8\\
        Act-One~\cite{runwayactone} & 29.84 & 0.817 & 25.07 & 0.259 & 135.2\\
        \midrule
        Ours & \textbf{25.70} & \textbf{0.823} & \textbf{28.44} & \textbf{0.238} & \textbf{110.3}\\
    \bottomrule
  \end{tabular}
  \caption{Quantitative comparisons with portrait animation methods on our collected dataset.}
  \label{tab:ours_port}
\vspace{-0.3cm}
\end{table}

\begin{table}
  \centering
  \footnotesize
  \begin{tabular}{l c c c c c c c c c}
    \toprule
     & FID$\downarrow$ & SSIM$\uparrow$ & PSNR$\uparrow$ & LPIPS$\downarrow$ & FVD$\downarrow$
    \\
        \midrule
        Single-R  & 28.22 & 0.798 & 25.86 & 0.223 & 120.5\\
        Multi-R (pseudo) & 26.53 & 0.812 & 26.22 & 0.219 & 116.6\\
    \bottomrule
  \end{tabular}
  \caption{Ablation study on multi-reference.}
  \label{tab:ablation}
\vspace{-0.1cm}
\end{table}

\subsection{Ablation Study}\label{sec:4.3}
We conducted comprehensive ablation studies to evaluate the impact of several core components of our method.

\begin{figure}
   \centering
   \includegraphics[width=0.99\linewidth]{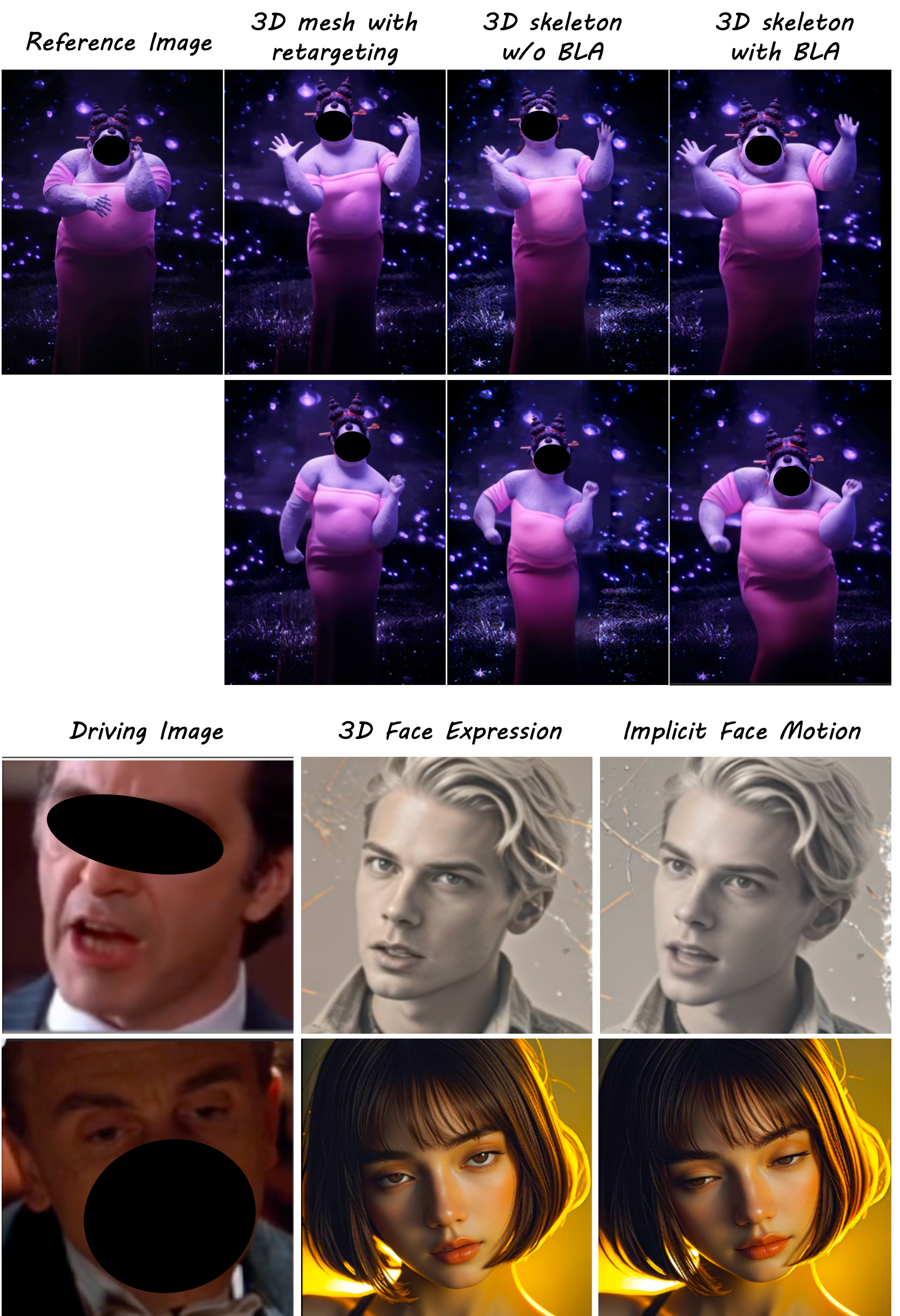}
   \caption{Ablation study of 3D skeletons with bone length adjustment (BLA) and implicit face features.}
   \label{fig:5-ablation}
\vspace{-0.3cm}
\end{figure}

\noindent\textbf{Multi-Reference Protocol.}
We compare two settings: (a) inference with a single reference image, (b) a two-stage inference approach as described in \cref{sec:3.3}, where pseudo reference images are generated first, followed by multi-reference inference. Results are shown in \cref{tab:ablation}. It demonstrates that pseudo multi-reference inference outperforms single-reference inference in terms of long-term video generation quality and temporal consistency. This is because during the extended video generation process, the supplementary reference images provide additional visual information about unseen areas, enabling the video generation process to leverage reference details. This helps avoid information loss and thereby maintains consistency throughout the video. Nevertheless, the performance achieved by a single reference image remains competitive, demonstrating its sufficiency for most scenarios.

\noindent\textbf{Hybrid Control Signals.}
We further investigate the contribution of our hybrid control signals by ablating key components: (a) substituting 3D head sphere and skeleton with 3D mesh (b) substituting implicit facial representations with 3d facial landmarks. Results are shown in \cref{fig:5-ablation}. It reveal a significant performance degradation under these settings, emphasizing the importance of each component in our hybrid control framework. Specifically, the 3D skeletons with bone length adjustment provide more accurate spatial guidance, and the implicit facial representations capture subtle expression details more effectively than traditional landmarks. These findings demonstrate the effectiveness and superiority of our proposed hybrid control signals in achieving high-quality and realistic human image animation.


\section{Conclusion}

In this paper, we present a holistic human image animation framework DreamActor-M1 addressing multi-scale adaptation, fine-grained facial expression and body movement control, and long-term consistency in unseen regions. We employ a progressive training strategy using data with varying resolutions and scales to handle various image scales ranging from portraits to full-body views.
By decoupling identity, body pose, and facial expression through hybrid control signals, our method achieves precise facial dynamics and vivid body movements while preserving the character identity. The proposed complementary appearance guidance resolves information gaps in cross-scale animation and unseen region synthesis. We believe these innovations provide potential insights for future research in complex motion modeling and real-world deployment of expressive human animation.

\noindent\textbf{Limitation.} Our framework faces inherent difficulties in controlling dynamic camera movements, and fails to generate physical interactions with environmental objects. In addition, the bone length adjustment of our method using~\cite{shi2024seededit} exhibits instability in edge cases, requiring multiple iterations with manual selections for optimal cases. 
These challenges still need to be addressed in future research.

\noindent\textbf{Ethics considerations.} 
Human image animation has possible social risks, like being misused to make fake videos. The proposed technology could be used to create fake videos of people, but existing detection tools~\cite{zhou2024dormant,chen2024demamba} can spot these fakes. To reduce these risks, clear ethical rules and responsible usage guidelines are necessary.
We will strictly restrict access to our core models and codes to prevent misuse. Images and videos are all from publicly available sources. If there are any concerns, please contact us and we will delete it in time.
\section*{Acknowledgment}

We extend our sincere gratitude to Shanchuan Lin, Lu Jiang, Zhurong Xia, Jianwen Jiang, Zerong Zheng, Chao Liang, Youjiang Xu, Ming Zhou, Tongchun Zuo, Xin Dong and Yanbo Zheng for their invaluable contributions and supports to this research work.

{
    \small
    \bibliographystyle{ieeenat_fullname}
    \bibliography{main}

\begin{thebibliography}{63}
\providecommand{\natexlab}[1]{#1}
\providecommand{\url}[1]{\texttt{#1}}
\expandafter\ifx\csname urlstyle\endcsname\relax
  \providecommand{\doi}[1]{doi: #1}\else
  \providecommand{\doi}{doi: \begingroup \urlstyle{rm}\Url}\fi

\bibitem[Chang et~al.(2023)Chang, Shi, Gao, Fu, Xu, Song, Yan, Zhu, Yang, and Soleymani]{chang2023magicpose}
Di Chang, Yichun Shi, Quankai Gao, Jessica Fu, Hongyi Xu, Guoxian Song, Qing Yan, Yizhe Zhu, Xiao Yang, and Mohammad Soleymani.
\newblock Magicpose: Realistic human poses and facial expressions retargeting with identity-aware diffusion.
\newblock \emph{arXiv preprint arXiv:2311.12052}, 2023.

\bibitem[Chen et~al.(2024)Chen, Hong, Huang, Xu, Gu, Li, Lan, Zhu, Zhang, Wang, et~al.]{chen2024demamba}
Haoxing Chen, Yan Hong, Zizheng Huang, Zhuoer Xu, Zhangxuan Gu, Yaohui Li, Jun Lan, Huijia Zhu, Jianfu Zhang, Weiqiang Wang, et~al.
\newblock Demamba: Ai-generated video detection on million-scale genvideo benchmark.
\newblock \emph{arXiv preprint arXiv:2405.19707}, 2024.

\bibitem[Deng et~al.(2024)Deng, Wang, Ren, Chen, and Wang]{deng2024portrait4d}
Yu Deng, Duomin Wang, Xiaohang Ren, Xingyu Chen, and Baoyuan Wang.
\newblock Portrait4d: Learning one-shot 4d head avatar synthesis using synthetic data.
\newblock In \emph{Proceedings of the IEEE/CVF Conference on Computer Vision and Pattern Recognition}, pages 7119--7130, 2024.

\bibitem[Doukas et~al.(2021)Doukas, Zafeiriou, and Sharmanska]{Doukas_2021_ICCV}
Michail~Christos Doukas, Stefanos Zafeiriou, and Viktoriia Sharmanska.
\newblock Headgan: One-shot neural head synthesis and editing.
\newblock In \emph{Proceedings of the IEEE/CVF International Conference on Computer Vision (ICCV)}, pages 14398--14407, 2021.

\bibitem[Drobyshev et~al.(2022)Drobyshev, Chelishev, Khakhulin, Ivakhnenko, Lempitsky, and Zakharov]{drobyshev2022megaportraits}
Nikita Drobyshev, Jenya Chelishev, Taras Khakhulin, Aleksei Ivakhnenko, Victor Lempitsky, and Egor Zakharov.
\newblock Megaportraits: One-shot megapixel neural head avatars.
\newblock In \emph{Proceedings of the 30th ACM International Conference on Multimedia}, pages 2663--2671, 2022.

\bibitem[Drobyshev et~al.(2024)Drobyshev, Casademunt, Vougioukas, Landgraf, Petridis, and Pantic]{drobyshev2024emoportraits}
Nikita Drobyshev, Antoni~Bigata Casademunt, Konstantinos Vougioukas, Zoe Landgraf, Stavros Petridis, and Maja Pantic.
\newblock Emoportraits: Emotion-enhanced multimodal one-shot head avatars.
\newblock In \emph{Proceedings of the IEEE/CVF Conference on Computer Vision and Pattern Recognition}, pages 8498--8507, 2024.

\bibitem[Esser et~al.(2024)Esser, Kulal, Blattmann, Entezari, M{\"u}ller, Saini, Levi, Lorenz, Sauer, Boesel, et~al.]{esser2024scaling}
Patrick Esser, Sumith Kulal, Andreas Blattmann, Rahim Entezari, Jonas M{\"u}ller, Harry Saini, Yam Levi, Dominik Lorenz, Axel Sauer, Frederic Boesel, et~al.
\newblock Scaling rectified flow transformers for high-resolution image synthesis.
\newblock In \emph{Forty-first international conference on machine learning}, 2024.

\bibitem[Gan et~al.(2025)Gan, Ren, Zhang, Ye, Xie, Yin, Yuan, Peng, and Zhu]{gan2025humandit}
Qijun Gan, Yi Ren, Chen Zhang, Zhenhui Ye, Pan Xie, Xiang Yin, Zehuan Yuan, Bingyue Peng, and Jianke Zhu.
\newblock Humandit: Pose-guided diffusion transformer for long-form human motion video generation.
\newblock \emph{arXiv preprint arXiv:2502.04847}, 2025.

\bibitem[Goel et~al.(2023)Goel, Pavlakos, Rajasegaran, Kanazawa, and Malik]{goel2023humans}
Shubham Goel, Georgios Pavlakos, Jathushan Rajasegaran, Angjoo Kanazawa, and Jitendra Malik.
\newblock Humans in 4d: Reconstructing and tracking humans with transformers.
\newblock In \emph{Proceedings of the IEEE/CVF International Conference on Computer Vision}, pages 14783--14794, 2023.

\bibitem[Guan et~al.(2024)Guan, Yang, Wang, Zhou, He, Xu, Feng, Ding, Wang, Xie, et~al.]{guan2024talk}
Jiazhi Guan, Quanwei Yang, Kaisiyuan Wang, Hang Zhou, Shengyi He, Zhiliang Xu, Haocheng Feng, Errui Ding, Jingdong Wang, Hongtao Xie, et~al.
\newblock Talk-act: Enhance textural-awareness for 2d speaking avatar reenactment with diffusion model.
\newblock In \emph{SIGGRAPH Asia 2024 Conference Papers}, pages 1--11, 2024.

\bibitem[G{\"u}ler et~al.(2018)G{\"u}ler, Neverova, and Kokkinos]{guler2018densepose}
R{\i}za~Alp G{\"u}ler, Natalia Neverova, and Iasonas Kokkinos.
\newblock Densepose: Dense human pose estimation in the wild.
\newblock In \emph{Proceedings of the IEEE conference on computer vision and pattern recognition}, pages 7297--7306, 2018.

\bibitem[Guo et~al.(2024)Guo, Zhang, Liu, Zhong, Zhang, Wan, and Zhang]{guo2024liveportrait}
Jianzhu Guo, Dingyun Zhang, Xiaoqiang Liu, Zhizhou Zhong, Yuan Zhang, Pengfei Wan, and Di Zhang.
\newblock Liveportrait: Efficient portrait animation with stitching and retargeting control.
\newblock \emph{arXiv preprint arXiv:2407.03168}, 2024.

\bibitem[Hong et~al.(2024)Hong, Xu, Liu, Lin, Song, Shu, Zhou, Ceylan, and Xu]{hong2024free}
Fa-Ting Hong, Zhan Xu, Haiyang Liu, Qinjie Lin, Luchuan Song, Zhixin Shu, Yang Zhou, Duygu Ceylan, and Dan Xu.
\newblock Free-viewpoint human animation with pose-correlated reference selection.
\newblock \emph{arXiv preprint arXiv:2412.17290}, 2024.

\bibitem[Hu(2024)]{hu2024animate}
Li Hu.
\newblock Animate anyone: Consistent and controllable image-to-video synthesis for character animation.
\newblock In \emph{Proceedings of the IEEE/CVF Conference on Computer Vision and Pattern Recognition}, pages 8153--8163, 2024.

\bibitem[Hu et~al.(2025)Hu, Wang, Shen, Gao, Meng, Zhuo, Zhang, Zhang, and Bo]{hu2025animate}
Li Hu, Guangyuan Wang, Zhen Shen, Xin Gao, Dechao Meng, Lian Zhuo, Peng Zhang, Bang Zhang, and Liefeng Bo.
\newblock Animate anyone 2: High-fidelity character image animation with environment affordance.
\newblock \emph{arXiv preprint arXiv:2502.06145}, 2025.

\bibitem[Huang et~al.(2024)Huang, Tang, Zhang, Cun, Cao, Li, and Lee]{huang2024make}
Ziyao Huang, Fan Tang, Yong Zhang, Xiaodong Cun, Juan Cao, Jintao Li, and Tong-Yee Lee.
\newblock Make-your-anchor: A diffusion-based 2d avatar generation framework.
\newblock In \emph{Proceedings of the IEEE/CVF Conference on Computer Vision and Pattern Recognition}, pages 6997--7006, 2024.

\bibitem[Jiang et~al.(2023)Jiang, Lu, Zhang, Ma, Han, Lyu, Li, and Chen]{jiang2023rtmpose}
Tao Jiang, Peng Lu, Li Zhang, Ningsheng Ma, Rui Han, Chengqi Lyu, Yining Li, and Kai Chen.
\newblock Rtmpose: Real-time multi-person pose estimation based on mmpose.
\newblock \emph{arXiv preprint arXiv:2303.07399}, 2023.

\bibitem[Karras et~al.(2023)Karras, Holynski, Wang, and Kemelmacher-Shlizerman]{karras2023dreampose}
Johanna Karras, Aleksander Holynski, Ting-Chun Wang, and Ira Kemelmacher-Shlizerman.
\newblock Dreampose: Fashion image-to-video synthesis via stable diffusion.
\newblock In \emph{2023 IEEE/CVF International Conference on Computer Vision (ICCV)}, pages 22623--22633. IEEE, 2023.

\bibitem[Kirschstein et~al.(2023)Kirschstein, Qian, Giebenhain, Walter, and Nie{\ss}ner]{kirschstein2023nersemble}
Tobias Kirschstein, Shenhan Qian, Simon Giebenhain, Tim Walter, and Matthias Nie{\ss}ner.
\newblock Nersemble: Multi-view radiance field reconstruction of human heads.
\newblock \emph{ACM Transactions on Graphics (TOG)}, 42\penalty0 (4):\penalty0 1--14, 2023.

\bibitem[Li et~al.(2024)Li, Li, Yang, Cao, Zhu, Cheng, and Chen]{li2024dispose}
Hongxiang Li, Yaowei Li, Yuhang Yang, Junjie Cao, Zhihong Zhu, Xuxin Cheng, and Long Chen.
\newblock Dispose: Disentangling pose guidance for controllable human image animation.
\newblock \emph{arXiv preprint arXiv:2412.09349}, 2024.

\bibitem[Lin et~al.(2025{\natexlab{a}})Lin, Jiang, Yang, Zheng, and Liang]{lin2025omnihuman}
Gaojie Lin, Jianwen Jiang, Jiaqi Yang, Zerong Zheng, and Chao Liang.
\newblock Omnihuman-1: Rethinking the scaling-up of one-stage conditioned human animation models.
\newblock \emph{arXiv preprint arXiv:2502.01061}, 2025{\natexlab{a}}.

\bibitem[Lin et~al.(2025{\natexlab{b}})Lin, Jiang, Yang, Zheng, and Liang]{lin2025omnihuman1}
Gaojie Lin, Jianwen Jiang, Jiaqi Yang, Zerong Zheng, and Chao Liang.
\newblock Omnihuman-1: Rethinking the scaling-up of one-stage conditioned human animation models.
\newblock \emph{arXiv preprint arXiv:2502.01061}, 2025{\natexlab{b}}.

\bibitem[Lin et~al.(2025{\natexlab{c}})Lin, Xia, Ren, Yang, Xiao, and Jiang]{lin2025diffusion}
Shanchuan Lin, Xin Xia, Yuxi Ren, Ceyuan Yang, Xuefeng Xiao, and Lu Jiang.
\newblock Diffusion adversarial post-training for one-step video generation.
\newblock \emph{arXiv preprint arXiv:2501.08316}, 2025{\natexlab{c}}.

\bibitem[Lipman et~al.(2022)Lipman, Chen, Ben-Hamu, Nickel, and Le]{lipman2022flow}
Yaron Lipman, Ricky~TQ Chen, Heli Ben-Hamu, Maximilian Nickel, and Matt Le.
\newblock Flow matching for generative modeling.
\newblock \emph{arXiv preprint arXiv:2210.02747}, 2022.

\bibitem[Ma et~al.(2024)Ma, Liu, Wang, Pan, He, Yuan, Zeng, Cai, Shum, Liu, et~al.]{ma2024followemoji}
Yue Ma, Hongyu Liu, Hongfa Wang, Heng Pan, Yingqing He, Junkun Yuan, Ailing Zeng, Chengfei Cai, Heung-Yeung Shum, Wei Liu, et~al.
\newblock Follow-your-emoji: Fine-controllable and expressive freestyle portrait animation.
\newblock In \emph{SIGGRAPH Asia 2024 Conference Papers}, pages 1--12, 2024.

\bibitem[Men et~al.(2024)Men, Yao, Cui, and Bo]{men2024mimo}
Yifang Men, Yuan Yao, Miaomiao Cui, and Liefeng Bo.
\newblock Mimo: Controllable character video synthesis with spatial decomposed modeling.
\newblock \emph{arXiv preprint arXiv:2409.16160}, 2024.

\bibitem[Pavlakos et~al.(2019)Pavlakos, Choutas, Ghorbani, Bolkart, Osman, Tzionas, and Black]{SMPL-X:2019}
Georgios Pavlakos, Vasileios Choutas, Nima Ghorbani, Timo Bolkart, Ahmed A.~A. Osman, Dimitrios Tzionas, and Michael~J. Black.
\newblock Expressive body capture: 3d hands, face, and body from a single image.
\newblock In \emph{Proceedings IEEE Conf. on Computer Vision and Pattern Recognition (CVPR)}, 2019.

\bibitem[Pavlakos et~al.(2024)Pavlakos, Shan, Radosavovic, Kanazawa, Fouhey, and Malik]{pavlakos2024reconstructing}
Georgios Pavlakos, Dandan Shan, Ilija Radosavovic, Angjoo Kanazawa, David Fouhey, and Jitendra Malik.
\newblock Reconstructing hands in 3{D} with transformers.
\newblock In \emph{CVPR}, 2024.

\bibitem[Peebles and Xie(2023)]{peebles2023scalable}
William Peebles and Saining Xie.
\newblock Scalable diffusion models with transformers.
\newblock In \emph{Proceedings of the IEEE/CVF international conference on computer vision}, pages 4195--4205, 2023.

\bibitem[Peng et~al.(2024)Peng, Wang, Zhang, Li, Yang, and Jia]{peng2024controlnext}
Bohao Peng, Jian Wang, Yuechen Zhang, Wenbo Li, Ming-Chang Yang, and Jiaya Jia.
\newblock Controlnext: Powerful and efficient control for image and video generation.
\newblock \emph{arXiv preprint arXiv:2408.06070}, 2024.

\bibitem[Qiu et~al.(2025)Qiu, Fei, Wang, Bai, Yu, Fan, Chen, and Wen]{qiu2025skyreels}
Di Qiu, Zhengcong Fei, Rui Wang, Jialin Bai, Changqian Yu, Mingyuan Fan, Guibin Chen, and Xiang Wen.
\newblock Skyreels-a1: Expressive portrait animation in video diffusion transformers.
\newblock \emph{arXiv preprint arXiv:2502.10841}, 2025.

\bibitem[Ren et~al.(2021)Ren, Li, Chen, Li, and Liu]{ren2021pirenderer}
Yurui Ren, Ge Li, Yuanqi Chen, Thomas~H. Li, and Shan Liu.
\newblock Pirenderer: Controllable portrait image generation via semantic neural rendering, 2021.

\bibitem[Rombach et~al.(2022)Rombach, Blattmann, Lorenz, Esser, and Ommer]{rombach2022high}
Robin Rombach, Andreas Blattmann, Dominik Lorenz, Patrick Esser, and Bj{\"o}rn Ommer.
\newblock High-resolution image synthesis with latent diffusion models.
\newblock In \emph{Proceedings of the IEEE/CVF conference on computer vision and pattern recognition}, pages 10684--10695, 2022.

\bibitem[Runway(2024)]{runwayactone}
Runway.
\newblock \url{https://runwayml.com/research/introducing-act-one}, 2024.

\bibitem[Shao et~al.(2024)Shao, Pang, Zheng, Sun, and Liu]{shao2024human4dit}
Ruizhi Shao, Youxin Pang, Zerong Zheng, Jingxiang Sun, and Yebin Liu.
\newblock Human4dit: 360-degree human video generation with 4d diffusion transformer.
\newblock \emph{arXiv preprint arXiv:2405.17405}, 2024.

\bibitem[Shi et~al.(2024)Shi, Wang, and Huang]{shi2024seededit}
Yichun Shi, Peng Wang, and Weilin Huang.
\newblock Seededit: Align image re-generation to image editing.
\newblock \emph{arXiv preprint arXiv:2411.06686}, 2024.

\bibitem[Siarohin et~al.(2019)Siarohin, Lathuili{\`e}re, Tulyakov, Ricci, and Sebe]{siarohin2019first}
Aliaksandr Siarohin, St{\'e}phane Lathuili{\`e}re, Sergey Tulyakov, Elisa Ricci, and Nicu Sebe.
\newblock First order motion model for image animation.
\newblock \emph{Advances in neural information processing systems}, 32, 2019.

\bibitem[Siarohin et~al.(2021)Siarohin, Woodford, Ren, Chai, and Tulyakov]{siarohin2021motion}
Aliaksandr Siarohin, Oliver~J Woodford, Jian Ren, Menglei Chai, and Sergey Tulyakov.
\newblock Motion representations for articulated animation.
\newblock In \emph{Proceedings of the IEEE/CVF conference on computer vision and pattern recognition}, pages 13653--13662, 2021.

\bibitem[Sun et~al.(2023)Sun, Wang, Wang, Li, Zhang, Zhang, and Liu]{sun2023next3d}
Jingxiang Sun, Xuan Wang, Lizhen Wang, Xiaoyu Li, Yong Zhang, Hongwen Zhang, and Yebin Liu.
\newblock Next3d: Generative neural texture rasterization for 3d-aware head avatars.
\newblock In \emph{CVPR}, 2023.

\bibitem[Tan et~al.(2025)Tan, Gong, Wang, Zhang, Zheng, Zheng, Zheng, Chen, and Yang]{AnimateX2025}
Shuai Tan, Biao Gong, Xiang Wang, Shiwei Zhang, Dandan Zheng, Ruobing Zheng, Kecheng Zheng, Jingdong Chen, and Ming Yang.
\newblock Animate-x: Universal character image animation with enhanced motion representation.
\newblock \emph{ICLR 2025}, 2025.

\bibitem[Tian et~al.(2024)Tian, Wang, Zhang, and Bo]{tian2024emo}
Linrui Tian, Qi Wang, Bang Zhang, and Liefeng Bo.
\newblock Emo: Emote portrait alive - generating expressive portrait videos with audio2video diffusion model under weak conditions, 2024.

\bibitem[Tong et~al.(2024)Tong, Li, Chen, Wu, and Zhou]{musepose}
Zhengyan Tong, Chao Li, Zhaokang Chen, Bin Wu, and Wenjiang Zhou.
\newblock Musepose: a pose-driven image-to-video framework for virtual human generation.
\newblock \emph{arxiv}, 2024.

\bibitem[Tu et~al.(2024)Tu, Xing, Han, Cheng, Dai, Luo, and Wu]{tu2024stableanimator}
Shuyuan Tu, Zhen Xing, Xintong Han, Zhi-Qi Cheng, Qi Dai, Chong Luo, and Zuxuan Wu.
\newblock Stableanimator: High-quality identity-preserving human image animation.
\newblock \emph{arXiv preprint arXiv:2411.17697}, 2024.

\bibitem[Wang et~al.(2023{\natexlab{a}})Wang, Deng, Yin, Shum, and Wang]{wang2022pdfgc}
Duomin Wang, Yu Deng, Zixin Yin, Heung-Yeung Shum, and Baoyuan Wang.
\newblock Progressive disentangled representation learning for fine-grained controllable talking head synthesis.
\newblock In \emph{Proceedings of the IEEE/CVF Conference on Computer Vision and Pattern Recognition (CVPR)}, 2023{\natexlab{a}}.

\bibitem[Wang et~al.(2022)Wang, Chen, Yu, Ma, Li, and Liu]{wang2022faceverse}
Lizhen Wang, Zhiyuan Chen, Tao Yu, Chenguang Ma, Liang Li, and Yebin Liu.
\newblock Faceverse: a fine-grained and detail-controllable 3d face morphable model from a hybrid dataset.
\newblock In \emph{Proceedings of the IEEE/CVF conference on computer vision and pattern recognition}, pages 20333--20342, 2022.

\bibitem[Wang et~al.(2023{\natexlab{b}})Wang, Zhang, Zhang, Gu, Bao, Baltrusaitis, Shen, Chen, Wen, Chen, et~al.]{wang2023rodin}
Tengfei Wang, Bo Zhang, Ting Zhang, Shuyang Gu, Jianmin Bao, Tadas Baltrusaitis, Jingjing Shen, Dong Chen, Fang Wen, Qifeng Chen, et~al.
\newblock Rodin: A generative model for sculpting 3d digital avatars using diffusion.
\newblock In \emph{Proceedings of the IEEE/CVF conference on computer vision and pattern recognition}, pages 4563--4573, 2023{\natexlab{b}}.

\bibitem[Wang et~al.(2024{\natexlab{a}})Wang, Li, Lin, Zhai, Lin, Yang, Zhang, Liu, and Wang]{wang2024disco}
Tan Wang, Linjie Li, Kevin Lin, Yuanhao Zhai, Chung-Ching Lin, Zhengyuan Yang, Hanwang Zhang, Zicheng Liu, and Lijuan Wang.
\newblock Disco: Disentangled control for realistic human dance generation.
\newblock In \emph{Proceedings of the IEEE/CVF Conference on Computer Vision and Pattern Recognition}, pages 9326--9336, 2024{\natexlab{a}}.

\bibitem[Wang et~al.(2021)Wang, Mallya, and Liu]{wang2021one}
Ting-Chun Wang, Arun Mallya, and Ming-Yu Liu.
\newblock One-shot free-view neural talking-head synthesis for video conferencing.
\newblock In \emph{Proceedings of the IEEE/CVF conference on computer vision and pattern recognition}, pages 10039--10049, 2021.

\bibitem[Wang et~al.(2024{\natexlab{b}})Wang, Zhang, Gao, Wang, Zhou, Zhang, Yan, and Sang]{wang2024unianimate}
Xiang Wang, Shiwei Zhang, Changxin Gao, Jiayu Wang, Xiaoqiang Zhou, Yingya Zhang, Luxin Yan, and Nong Sang.
\newblock Unianimate: Taming unified video diffusion models for consistent human image animation.
\newblock \emph{arXiv preprint arXiv:2406.01188}, 2024{\natexlab{b}}.

\bibitem[Xie et~al.(2024)Xie, Xu, Song, Wang, Shi, and Luo]{xie2024x}
You Xie, Hongyi Xu, Guoxian Song, Chao Wang, Yichun Shi, and Linjie Luo.
\newblock X-portrait: Expressive portrait animation with hierarchical motion attention.
\newblock In \emph{ACM SIGGRAPH 2024 Conference Papers}, pages 1--11, 2024.

\bibitem[Xu et~al.(2025)Xu, Chen, Guo, Yang, Li, Zang, Zhang, Tong, and Guo]{xu2025vasa}
Sicheng Xu, Guojun Chen, Yu-Xiao Guo, Jiaolong Yang, Chong Li, Zhenyu Zang, Yizhong Zhang, Xin Tong, and Baining Guo.
\newblock Vasa-1: Lifelike audio-driven talking faces generated in real time.
\newblock \emph{Advances in Neural Information Processing Systems}, 37:\penalty0 660--684, 2025.

\bibitem[Xu et~al.(2024{\natexlab{a}})Xu, Song, Song, Zhang, Liew, Xu, Xie, Luo, Lin, Feng, et~al.]{xu2024high}
Zhongcong Xu, Chaoyue Song, Guoxian Song, Jianfeng Zhang, Jun~Hao Liew, Hongyi Xu, You Xie, Linjie Luo, Guosheng Lin, Jiashi Feng, et~al.
\newblock High quality human image animation using regional supervision and motion blur condition.
\newblock \emph{arXiv preprint arXiv:2409.19580}, 2024{\natexlab{a}}.

\bibitem[Xu et~al.(2024{\natexlab{b}})Xu, Zhang, Liew, Yan, Liu, Zhang, Feng, and Shou]{xu2024magicanimate}
Zhongcong Xu, Jianfeng Zhang, Jun~Hao Liew, Hanshu Yan, Jia-Wei Liu, Chenxu Zhang, Jiashi Feng, and Mike~Zheng Shou.
\newblock Magicanimate: Temporally consistent human image animation using diffusion model.
\newblock In \emph{Proceedings of the IEEE/CVF Conference on Computer Vision and Pattern Recognition}, pages 1481--1490, 2024{\natexlab{b}}.

\bibitem[Yang et~al.(2024)Yang, Li, Wu, Jing, Li, Ji, Liang, Fan, and Wang]{yang2024megactor}
Shurong Yang, Huadong Li, Juhao Wu, Minhao Jing, Linze Li, Renhe Ji, Jiajun Liang, Haoqiang Fan, and Jin Wang.
\newblock Megactor-${\Sigma}$: Unlocking flexible mixed-modal control in portrait animation with diffusion transformer.
\newblock \emph{arXiv preprint arXiv:2408.14975}, 2024.

\bibitem[Yoon et~al.(2024)Yoon, Koo, Lee, and Yoo]{yoon2024tpc}
Sunjae Yoon, Gwanhyeong Koo, Younghwan Lee, and Chang Yoo.
\newblock Tpc: Test-time procrustes calibration for diffusion-based human image animation.
\newblock \emph{Advances in Neural Information Processing Systems}, 37:\penalty0 118654--118677, 2024.

\bibitem[Yu et~al.(2023{\natexlab{a}})Yu, Lezama, Gundavarapu, Versari, Sohn, Minnen, Cheng, Birodkar, Gupta, Gu, et~al.]{yu2023language}
Lijun Yu, Jos{\'e} Lezama, Nitesh~B Gundavarapu, Luca Versari, Kihyuk Sohn, David Minnen, Yong Cheng, Vighnesh Birodkar, Agrim Gupta, Xiuye Gu, et~al.
\newblock Language model beats diffusion--tokenizer is key to visual generation.
\newblock \emph{arXiv preprint arXiv:2310.05737}, 2023{\natexlab{a}}.

\bibitem[Yu et~al.(2023{\natexlab{b}})Yu, Fan, Zhang, Wang, Yin, Bai, Cao, Shan, Wu, Sun, et~al.]{yu2023nofa}
Wangbo Yu, Yanbo Fan, Yong Zhang, Xuan Wang, Fei Yin, Yunpeng Bai, Yan-Pei Cao, Ying Shan, Yang Wu, Zhongqian Sun, et~al.
\newblock Nofa: Nerf-based one-shot facial avatar reconstruction.
\newblock In \emph{ACM SIGGRAPH 2023 conference proceedings}, pages 1--12, 2023{\natexlab{b}}.

\bibitem[Zhang et~al.(2024)Zhang, Gu, Wang, Wang, Cheng, Zhu, and Zou]{zhang2024mimicmotion}
Yuang Zhang, Jiaxi Gu, Li-Wen Wang, Han Wang, Junqi Cheng, Yuefeng Zhu, and Fangyuan Zou.
\newblock Mimicmotion: High-quality human motion video generation with confidence-aware pose guidance.
\newblock \emph{arXiv preprint arXiv:2406.19680}, 2024.

\bibitem[Zhao and Zhang(2022)]{zhao2022thin}
Jian Zhao and Hui Zhang.
\newblock Thin-plate spline motion model for image animation.
\newblock In \emph{Proceedings of the IEEE/CVF Conference on Computer Vision and Pattern Recognition}, pages 3657--3666, 2022.

\bibitem[Zhao et~al.(2025)Zhao, Xu, Song, Xie, Zhang, Li, Luo, Suo, and Liu]{X-NeMo2025}
Xiaochen Zhao, Hongyi Xu, Guoxian Song, You Xie, Chenxu Zhang, Xiu Li, Linjie Luo, Jinli Suo, and Yebin Liu.
\newblock X-nemo: Expressive neural motion reenactment via disentangled latent attention.
\newblock \emph{ICLR 2025}, 2025.

\bibitem[Zhou et~al.(2024)Zhou, Wang, Li, Meng, and Chen]{zhou2024dormant}
Jiachen Zhou, Mingsi Wang, Tianlin Li, Guozhu Meng, and Kai Chen.
\newblock Dormant: Defending against pose-driven human image animation.
\newblock \emph{arXiv preprint arXiv:2409.14424}, 2024.

\bibitem[Zhu et~al.(2024{\natexlab{a}})Zhu, Chen, Dai, Dong, Xu, Cao, Yao, Zhu, and Zhu]{zhu2024champ}
Shenhao Zhu, Junming~Leo Chen, Zuozhuo Dai, Zilong Dong, Yinghui Xu, Xun Cao, Yao Yao, Hao Zhu, and Siyu Zhu.
\newblock Champ: Controllable and consistent human image animation with 3d parametric guidance.
\newblock In \emph{European Conference on Computer Vision}, pages 145--162. Springer, 2024{\natexlab{a}}.

\bibitem[Zhu et~al.(2024{\natexlab{b}})Zhu, Zhang, Rong, Hu, Liang, and Ge]{zhu2024infp}
Yongming Zhu, Longhao Zhang, Zhengkun Rong, Tianshu Hu, Shuang Liang, and Zhipeng Ge.
\newblock Infp: Audio-driven interactive head generation in dyadic conversations.
\newblock \emph{arXiv preprint arXiv:2412.04037}, 2024{\natexlab{b}}.

\end{thebibliography}
}

\end{document}